\documentclass{article}
\usepackage{spconf,amsmath,graphicx}
\usepackage{booktabs}
\usepackage{graphics}
\usepackage{hhline}
\usepackage{multirow}
\usepackage{graphicx}
\usepackage{amssymb}
\usepackage{amsthm,mwe,graphbox,fixltx2e}
\usepackage[export]{adjustbox}
\usepackage{amsmath,textcomp,threeparttable,gensymb}
\usepackage{color,url,rotating}
\input{boldcommands}
\usepackage[bookmarks=false,bookmarksopen=false,
breaklinks=true,colorlinks=true,linkcolor=blue,anchorcolor=blue,citecolor=blue,
filecolor=blue,menucolor=blue,urlcolor=blue]{hyperref}
\newcommand{\ts}{\textsuperscript}

\title{TBI Contusion Segmentation from MRI using Convolutional Neural Networks }

\name{Snehashis Roy{$^a$}
\sthanks{This work was supported by the Department of Defense in the Center for Neuroscience and 
Regenerative Medicine and the Intramural Research Program of the National Institutes of Health. 
This work was also partially supported by grant from National MS Society RG-1507-05243.}
\qquad John A. Butman{$^b$} \qquad Leighton Chan{$^c$} 
\qquad Dzung L. Pham{$^a$}}
\address{$^{a}$ Center for Neuroscience and Regenerative Medicine, Henry Jackson Foundation\\
$^{b}$ Radiology and Imaging Sciences, Clinical Center, National Institute of Health \\
$^{c}$ Rehabilitation Medicine Department, National Institute of Health
}

\begin{document}

\maketitle

\begin{abstract}
Traumatic brain injury (TBI) is caused by a sudden trauma to the head that may result in hematomas 
and contusions and can lead to stroke or chronic disability. An accurate quantification of the  
lesion volumes and their locations is essential to understand the pathophysiology of TBI and its 
progression. In this paper, we propose a fully convolutional neural network (CNN) model to segment 
contusions and lesions from brain magnetic resonance (MR) images of patients with TBI. The CNN 
architecture proposed here was based on a state of the art CNN architecture from Google, called 
Inception. Using a $3$-layer Inception network, lesions are segmented from multi-contrast MR images. 
When compared with two recent TBI lesion segmentation methods, one based on CNN (called DeepMedic) 
and another based on random forests, the proposed algorithm showed improved 
segmentation accuracy on images of $18$ patients with mild to severe TBI. Using a leave-one-out 
cross validation, the proposed model achieved a median Dice of $0.75$, which was significantly 
better ($p<0.01$) than the two competing methods.

\end{abstract}


\section{Introduction}
\label{sec:intro}
Accurate segmentation of contusions, edema, hemorrhages, and lesions is important for understanding 
the effects of traumatic brain injury (TBI). Although the exact pathophysiology of TBI is unknown, 
TBI lesions have been shown to be associated with cognitive decline \cite{kinnunen2011} and other 
neurological impairments \cite{trifan2017}. Quantitative information about types, locations, and 
volumes of the lesions can be correlated with the patient outcome and mortality \cite{moen2012} to 
better identify patients at risk for poor outcomes. The presence of TBI is typically assessed from 
CT images in a clinical setting. MR images provide details of TBI lesions that are not easily 
observed on CT because MR images have higher resolution and better soft tissue contrast.
Therefore segmentation of TBI lesions from magnetic resonance imaging (MRI) may provide valuable 
insights into the understanding of TBI.

\begin{figure}[!bt]
\begin{center}
\tabcolsep 0pt
\begin{tabular}{cccc}
\includegraphics[height=0.15\textwidth]{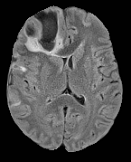}  &
\includegraphics[height=0.15\textwidth]{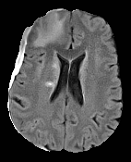}  &
\includegraphics[height=0.15\textwidth]{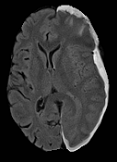}  &
\includegraphics[height=0.15\textwidth]{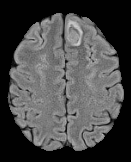}  
\end{tabular}
\end{center}
\vspace{-1em}
\caption{Examples of hemorrhages are shown on FLAIR images for $4$ patients with TBI.
}
\label{fig:examples}
\end{figure}

Many machine learning algorithms have been proposed for segmenting lesions from MR images
from patients with multiple sclerosis (MS) and Alzheimer's disease (AD). However, 
TBI contusions and lesions are more heterogeneous compared to MS or AD, making automated 
segmentation a challenging task. An example of different types of TBI lesions are shown in 
Fig.~\ref{fig:examples}. Clearly, there is considerable variability of the lesions in terms of 
shape, size, location, and overall intensity distribution. Instead of segmenting different parts of 
the lesions, such as encephalomalacia or blood, we propose to label all the abnormal tissue in 
a binary classification.

\begin{figure*}[!tbh]
\begin{center}
\includegraphics[width=1\textwidth]{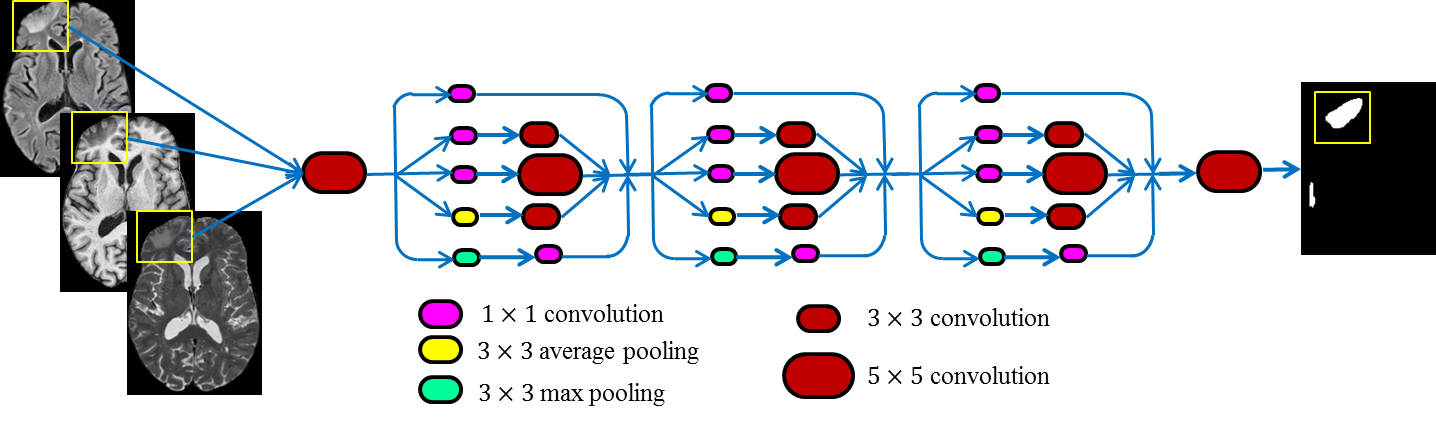}  
\end{center}
\vspace{-1em}
\caption{The proposed CNN architecture using modified Google Inception \cite{szegedy2015} block
is shown. Patches from MPRAGE, $T_2$-w and FLAIR images are passed through $3$ modified Inception
blocks, the output being a patch of lesion memberships.
}
\label{fig:network}
\end{figure*}

Most of the previously successful lesion segmentation methods are supervised; training data are 
derived from manual lesion delineations on multi-contrast MR images, and machine learning 
algorithms, such as random forest \cite{zikic2012,rao2014}, are used to learn a mapping between 
intensities and lesion labels. In atlas based methods, multiple normal atlases are registered to a 
subject brain with pathologies to detect voxels with out-of-atlas intensity distributions 
\cite{asman2013}, which correspond to pathological regions. Similarly, label fusion type techniques 
use tissue priors from normal atlases to detect voxels with pathological intensities and segment 
whole brain \cite{ledig2015}. Dictionary learning has also been applied for lesion segmentation 
where image patch dictionaries are learnt for patches containing normal tissues and lesions  
\cite{roy2015,roy2013}. 

Recently, convolutional neural networks (CNNs), or deep learning \cite{hinton2015}, have been 
successful in tumor and stroke lesion segmentation \cite{menze2015}. Unlike traditional machine
learning methods, CNNs do not need hand crafted features, which make them applicable to diverse
problems. In recent tumor segmentation challenges \cite{menze2015}, the top rated methods have
used neural networks. These methods employ either 2D \cite{havaei2017,ronnenberger2015} 
or 3D \cite{kamnitsas2017} image patches containing normal tissues as well as lesion voxels, and  
dense networks were trained using manual delineations of lesions. CNNs have the advantage of being 
fast because they are implemented in GPUs; segmentation of a subject brain can be completed in only 
couple minutes \cite{kamnitsas2017}. 

In this paper, we propose an automated TBI lesion segmentation method using CNNs. We explored a
recent state-of-the-art network architecture called Inception \cite{szegedy2015}, initially proposed 
to classify natural images by Google, to segment lesions using multi-contrast MR images. We compared 
this method with a random forest based method \cite{zikic2012} as well as another recent CNN based 
method DeepMedic \cite{kamnitsas2017} to show that the proposed method produced improved 
segmentation performance.

\section{Dataset}
The proposed segmentation was evaluated on multi-contrast images of $18$ patients with mild to
severe TBI. MPRAGE, $T_2$-w, and FLAIR images were acquired on a Siemens 3T scanners. The imaging
parameters were as follows, MPRAGE: $T_R=2530$ms, $T_E=3.03$ms, $T_I=1100$ms, flip angle $7^\circ$, 
resolution $1\times 1\times 1$ mm\ts{3}, size $256\times 256\times 176$; $T_2$: $T_R=3200$ms, 
$T_E=409$ms, flip angle $120^\circ$, resolution $0.98\times 0.98\times 1$ mm\ts{3}, size
$512\times 512\times 176$; FLAIR: $T_R=9.09$s, $T_E=112$ms, $T_I=2450$ms, flip angle $120^\circ$, 
resolution $0.86\times 0.86\times 3$ mm\ts{3}, size $256\times 256\times 50$. FLAIR and $T_2$-w
images were rigidly registered \cite{avants2011} to the MPRAGE. Then for every subject, the MPRAGE 
was stripped \cite{roy2016} and the same brainmask was applied to the other contrasts. All of the
stripped images were corrected for any intensity inhomogeneity \cite{tustison2010}. Finally TBI 
lesions were manually delineated on the registered FLAIR to serve as training and evaluation data. 
The manual segmentations were also examined by a neuroradiologist for accuracy.

\begin{figure*}[!tbh]
\begin{center}
\tabcolsep 0pt
\begin{tabular}{cccccccc}
& \textsc{MPRAGE} & \textsc{T2} & \textsc{FLAIR} & \textsc{RF} & \textsc{DeepMedic} & 
\textsc{Proposed} &  \textsc{Manual} \\
\rotatebox{90}{\hspace{2em}Subject \#1} &
\includegraphics[width=0.14\textwidth]{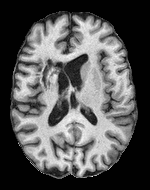}  &
\includegraphics[width=0.14\textwidth]{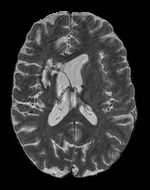}  &
\includegraphics[width=0.14\textwidth]{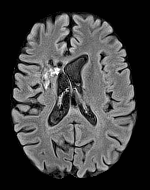}  &
\includegraphics[width=0.14\textwidth]{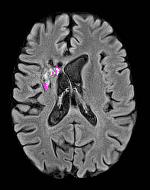}  &
\includegraphics[width=0.14\textwidth]{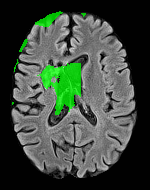}  &
\includegraphics[width=0.14\textwidth]{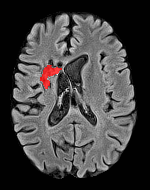}  &
\includegraphics[width=0.14\textwidth]{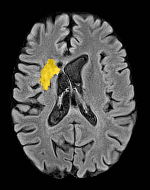} \\
\rotatebox{90}{\hspace{2em}Subject \#2} &
\includegraphics[width=0.14\textwidth]{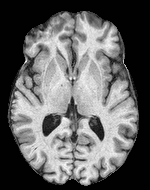}  &
\includegraphics[width=0.14\textwidth]{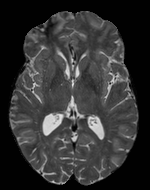}  &
\includegraphics[width=0.14\textwidth]{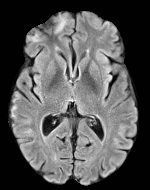}  &
\includegraphics[width=0.14\textwidth]{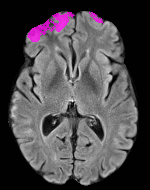}  &
\includegraphics[width=0.14\textwidth]{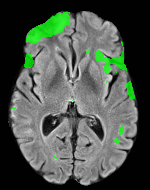}  &
\includegraphics[width=0.14\textwidth]{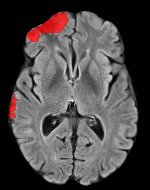}  &
\includegraphics[width=0.14\textwidth]{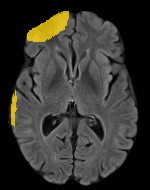} \\
\rotatebox{90}{\hspace{2em}Subject \#3} &
\includegraphics[width=0.14\textwidth]{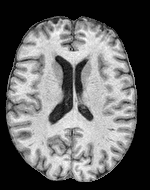}  &
\includegraphics[width=0.14\textwidth]{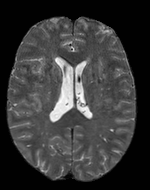}  &
\includegraphics[width=0.14\textwidth]{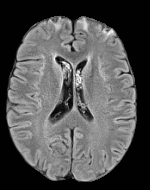}  &
\includegraphics[width=0.14\textwidth]{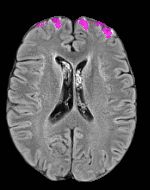}  &
\includegraphics[width=0.14\textwidth]{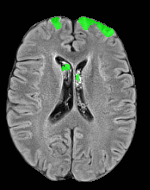}  &
\includegraphics[width=0.14\textwidth]{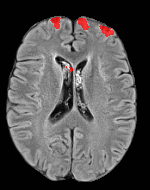}  &
\includegraphics[width=0.14\textwidth]{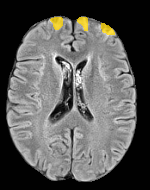} 
\end{tabular}
\end{center}
\vspace{-1em}
\caption{Examples of segmentations are shown for $3$ subjects. RF and DeepMedic corresponds
to \cite{zikic2012} and \cite{kamnitsas2017}, respectively.
}
\label{fig:results}
\end{figure*}

\section{Method}
The proposed fully convolutional network architecture is shown in Fig.~\ref{fig:network}. The 
network used cascades of $3$ modified Inception modules \cite{szegedy2015}. Each module consisted of 
parallel layers of convolutions, max-pooling, and average pooling. Each convolution was followed by 
a rectified linear unit (ReLU), which clamped any non-zero element in the convolution output to 
zero. Max-pooling and average pooling involve choosing a maximum value or average value over a 
sliding window of pre-defined size. The original Inception module, as proposed in 
\cite{szegedy2015}, consisted of $4$ parallel pathways, three $1\times 1$ convolutions and a 
max-pooling layer. We modified this architecture by adding another parallel pathway with an average 
pooling layer (yellow boxes in Fig.~\ref{fig:network}) followed 
by $3\times 3$ convolutions (see Fig.~\ref{fig:network}). The rationale behind adding this extra 
layer arose from the visual representation of the lesions, shown in Fig.~\ref{fig:examples}. 
Lesions in TBI can have different intensity profiles, both bright and dark depending on the time 
between the injury and the scan. To detect these intensities, average pooling acts as a low 
resolution feature map without actually having to downsample the image. This can have some 
practical implications which are explained later. The number of filters for each convolution was 
kept the same as the original paper \cite{szegedy2015}, with $32$ average pooling filters in each 
modified Inception module.

In recent famous CNN architectures such as ZFNet \cite{zeiler2014} or VGGNet \cite{simoyan2014}, 
convolutions and pooling were used in a serial manner and fully connected (or dense) layers were
added at the end of the networks for classification purpose. As a consequence, the number of free 
parameters to estimate increases exponentially with respect to layers; for example the number of 
parameters in VGGNet is 
$140\times 10^6$. This can cause optimization instability when the number of training examples is 
many fewer than the number of free parameters. The Inception module addresses this issue by using 
$1\times 1$ convolutions and having pooling and convolutions in parallel (see 
Fig.~\ref{fig:network}). Similar to ResNet \cite{he2016}, we excluded any fully connected layer.

During training, the training input consisted of mini-batches of $p\times p$ patch triplets, one for 
each MR contrast, from multiple atlases, where the center voxel of every patch has a non-zero label 
in the manual segmentations. A Gaussian blurred version of the manual segmentation, analogous to a 
membership, was used as the training output for estimating the parameters of the network. Therefore, 
for every training MR patch, instead of predicting the lesion label of the center voxel, we 
predict the $p\times p$ membership of the whole patch by using mean squared error as the metric. 
There are some advantages of using such a training procedure.
\begin{enumerate}
\item Since the network is fully convolutional and includes low-resolution features via average 
pooling instead of downsampling and upsampling, the training atlas image sizes do not need to be the same. 
Similarly, the testing image size also need not be the same as the training atlas sizes, as was the 
case in \cite{ronnenberger2015}. This has a practical advantage in that the trained model can be 
applied to a subject image of any size without knowing the atlas dimensions.
\item The total number of parameters is only approximately $294,000$, which is comparable to the 
number of training patches. 
\item With the removal of fully connected layers and direct estimation of memberships via minimizing 
mean squared error, we observed that the resultant memberships were crisper, thereby producing
fewer false positives than a comparable network with a fully connected layer, such as DeepMedic. 
\end{enumerate}

We used $45\times 45$ patches to train the network. Since the patches were 2D, we reoriented
the atlases into axial, coronal, and sagittal orientations, and trained one model for
each orientation separately. Adam \cite{kingma2015} was used to optimize the convolution filter 
weights via stochastic gradient descent. An average of 450,000 patches were used for training 
using $17$ atlases, of which $20$\% was used for validation. Training using $17$ atlases took about 
$10$ hours for $10$ epochs with a mini-batch size of $64$. Although the original Inception paper 
\cite{szegedy2015} proposed $11$ cascaded modules, limited experiments showed only $3$ modules to be
sufficient for an accurate segmentation. For a new subject, three memberships were generated 
for each orientation and then averaged to get the final membership. The memberships were thresholded 
at $0.5$ to obtain a binary segmentation. To decrease false positives, small disconnected objects
with volumes less than a certain threshold were removed from the binary segmentation. The volume 
threshold was computed as $90$\ts{th} percentile of volumes of all the $18$-connected objects in a
binary segmentation. Generating memberships on a new subject takes about $1$ minute.

\section{Results}
The proposed method was compared with a random forest based algorithm \cite{zikic2012}, denoted by
RF and implemented in-house, as well as another CNN based algorithm DeepMedic \cite{kamnitsas2017}. 
A leave-one-out cross-validation was performed on the $18$ subjects for each of the three competing 
methods. Example segmentations of the three methods are shown in Fig.~\ref{fig:results} for $3$ 
subjects. On average, RF resulted in under-segmentation and DeepMedic resulted in over-segmentation 
and more false positives. DeepMedic showed false positives near the right frontal lobe in subject 
\#1 and left temporal lobe in
subject \#2 , where there are some hyperintense artifacts on FLAIR. Neither RF nor the proposed 
method had any false positive around those artifacts. On subject \#1, the ventricles in FLAIR
were incorrectly segmented by DeepMedic. There were some CSF flow artifacts in the ventricles of 
subject \#3, which were segmented as lesions by both DeepMedic and the proposed method.

\begin{figure}[!bt]
\begin{center}
\tabcolsep 0pt
\begin{tabular}{cc}
\includegraphics[width=0.25\textwidth]{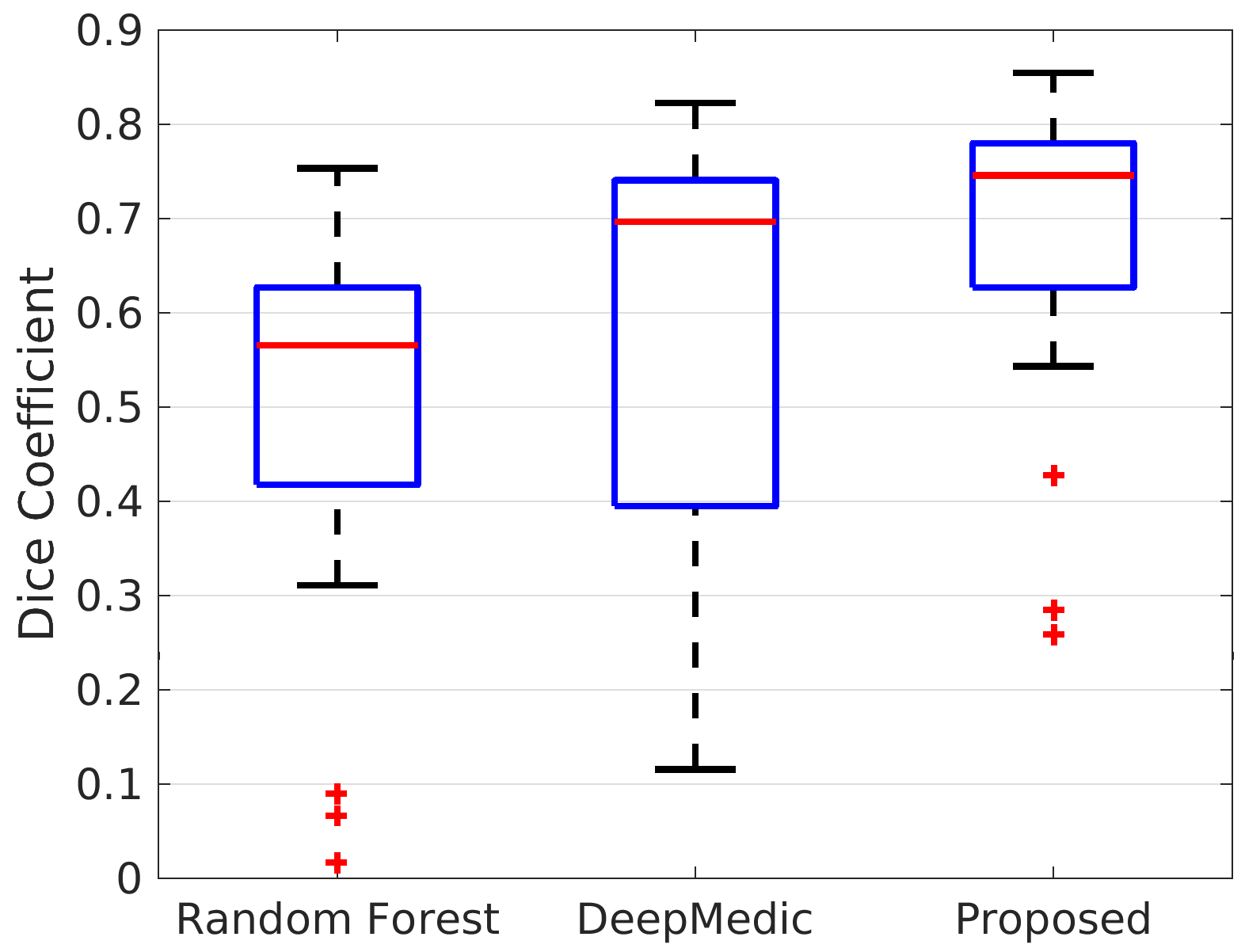}  &
\includegraphics[width=0.25\textwidth]{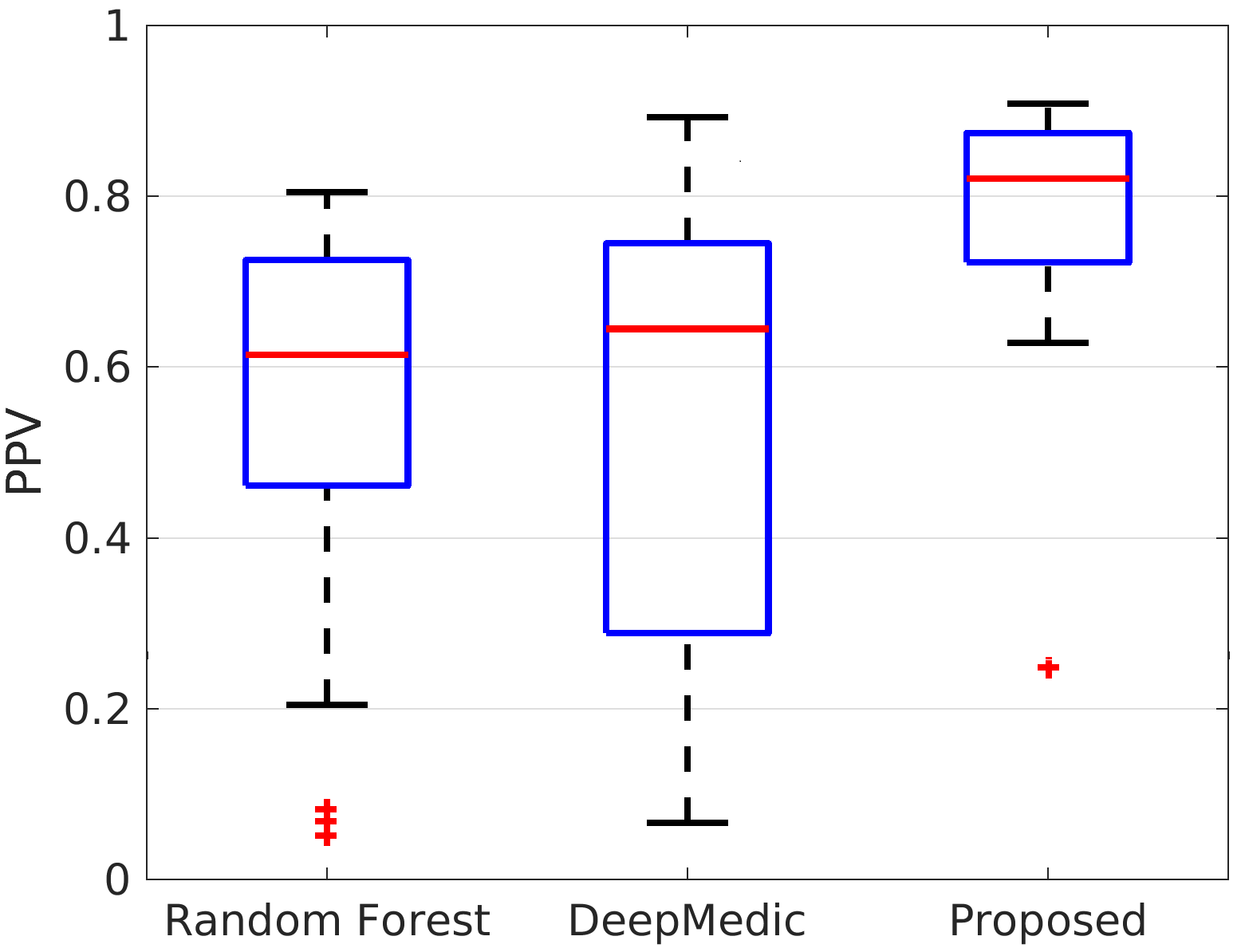}  \\
\includegraphics[width=0.25\textwidth]{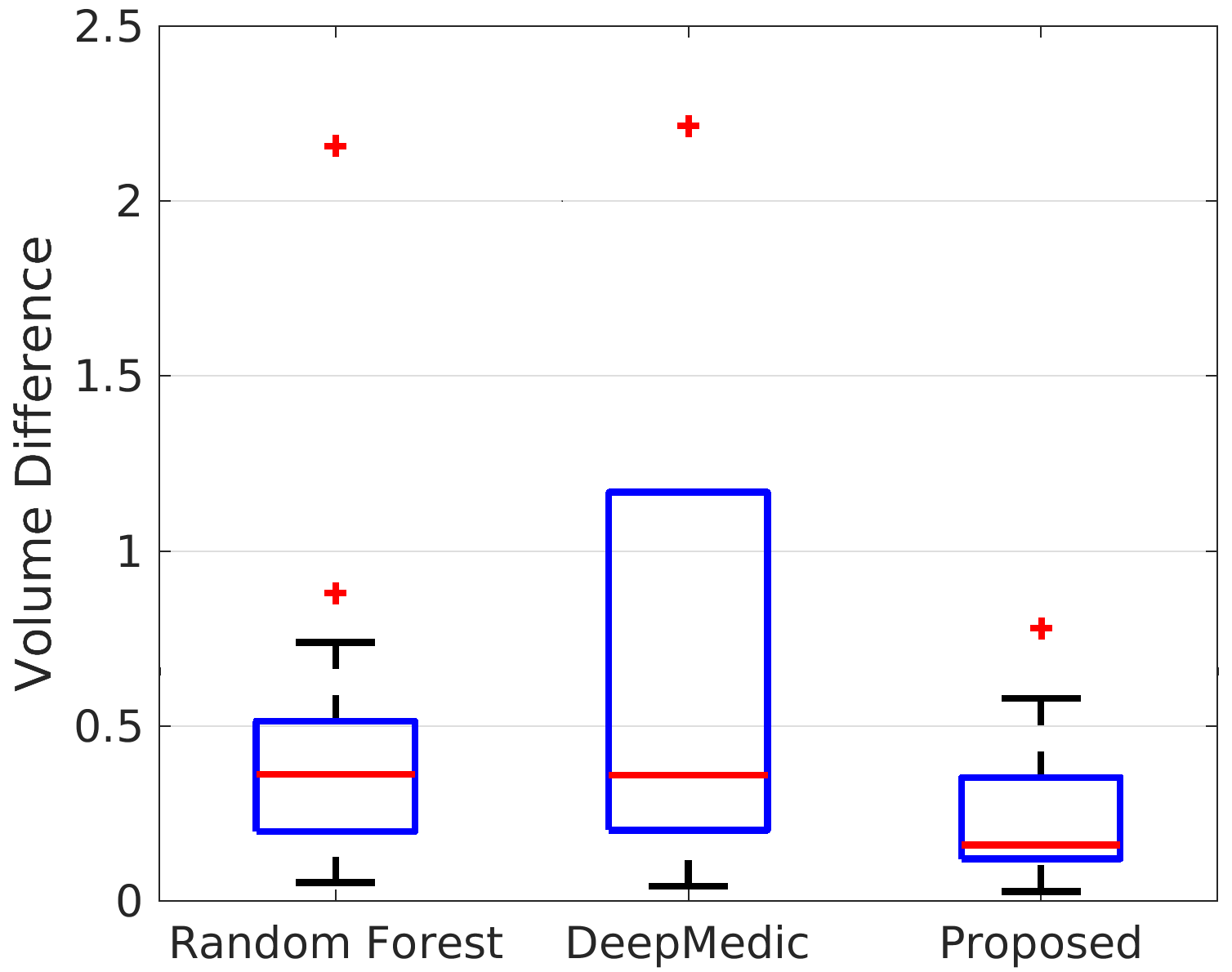}  &
\includegraphics[width=0.25\textwidth]{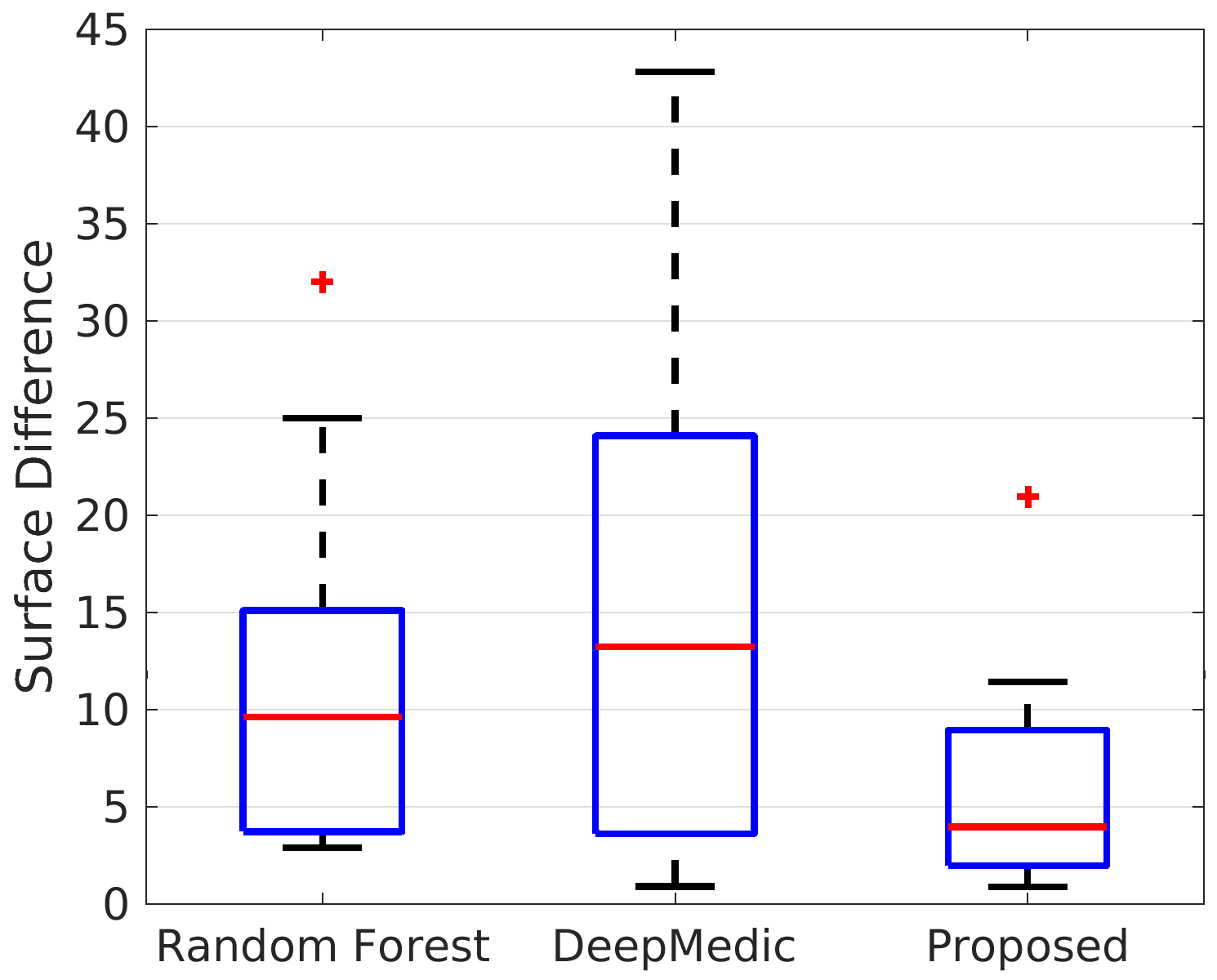}  
\end{tabular}
\end{center}
\vspace{-1em}
\caption{Quantitative comparison between the three competing methods are shown. The proposed method 
shows significant ($p<0.05$) improvement over other two methods for all four metrics.
}
\label{fig:plots}
\end{figure}

For quantitative comparison, we used Dice coefficient, positive predictive value (PPV), volume
difference (VD), and surface difference (SD). Assuming $\mathcal{A}$ and $\mathcal{M}$ to be the 
automated and the manual binary segmentations, Dice is defined as 
$\frac{2|\mathcal{A}\cap \mathcal{M}|}{|\mathcal{A}| + |\mathcal{M}|}$, where $|\cdot|$ indicates
number of non-zero voxels. PPV is defined as the ratio of true positives and total number of 
positive voxels, i.e., $\frac{2|\mathcal{A} \cap \mathcal{M}|}{|\mathcal{A}|}$. VD is defined
as the ratio between absolute volume difference and manual volume $\frac{abs(|\mathcal{A}| -  
|\mathcal{M}|)}{|\mathcal{M}|}$. SD is defined as the average of the Hausdorff distances from 
$\mathcal{M}$ to $\mathcal{A}$ and  $\mathcal{A}$ to $\mathcal{M}$.

Fig.~\ref{fig:plots} shows boxplots of Dice, PPV, VD, and SD of the three methods. Median
Dice coefficients for RF, DeepMedic, and our method were $0.566$, $0.697$, and $0.746$. Median
PPV values were $0.614$, $0.645$, and $0.821$, respectively. The proposed
method produced significantly higher Dice and PPV ($p<0.01$) and lower VD and SD ($p<0.05$)
compared to the other two methods. Median VD and SD values were $\{0.362,0.360,0.160\}$ and
$\{9.62,13.23,3.97\}$, respectively. DeepMedic showed higher Dice ($p=0.005$) but similar PPV, SD, 
and VD compared to RF ($p>0.25$). As observed in Fig.~\ref{fig:results}, DeepMedic had more false
positives than the proposed method, evident from its high VD and low PPV. We observed that most of 
the false positives in our method arose from flow artifacts inside the lateral and 3rd ventricles,
as shown in Fig.~\ref{fig:results} subject \#3. The false positives for RF and DeepMedic were also 
in the cortical region, especially in the temporal cortex, such as Fig.~\ref{fig:results} subject 
\#2.

\section{Discussion}
We have described a fully convolutional neural network based algorithm to segment TBI lesions from
multi-contrast MR images. Comparison with a random forest based algorithm and another CNN based
algorithm on $18$ subjects showed that our method produces more accurate segmentation and less false 
positives. 

We hypothesize that the low false positive rate in our proposed
method was attributed to bigger patch sizes ($45^2$ vs $17^3$ in DeepMedic) and removal of 
fully connected layers. Bigger patches provide more context to a lesion, thereby reducing the 
risk of false positives. This is seen in Fig.~\ref{fig:results} subject \#3, where the false
positives inside the ventricles was reduced in the proposed method.

We have used 2D patches separately in three orientations to train three models. The rationale
behind using 2D patches is that the FLAIR images, which usually produce the best contrast for
lesions, are anisotropic ($1\times 1\times 3$ mm\ts{3}) in resolution. So a full isotropic patch
may include spurious or irrelevant information regarding a small lesion. Also by using 2D models,
we were able to use large ($4.5 \times 4.5$ cm) patches which include more context to a small region
of interest, while fitting within a small GPU memory. Future work will include optimization of
the patch size and the depth of the network, comparison with full 3D patches, and differentiating
types of lesions.

\bibliographystyle{IEEEbib}
\small{
\bibliography{refs}
}

\end{document}